\documentclass[conference]{IEEEtran}
\IEEEoverridecommandlockouts
\usepackage{cite}
\usepackage{amsmath,amssymb,amsfonts}
\usepackage{algorithmic}
\usepackage{graphicx}
\usepackage{textcomp}
\usepackage{xcolor}
\usepackage[T1]{fontenc}
\usepackage{subcaption}
\usepackage{hyperref}
\def\BibTeX{{\rm B\kern-.05em{\sc i\kern-.025em b}\kern-.08em
    T\kern-.1667em\lower.7ex\hbox{E}\kern-.125emX}}
\begin{document}

\newcommand{\modelname}{P2P0.1}

\title{Pixels to Play: A Foundation Model for 3D Gameplay
\thanks{\textsuperscript{*} Short paper.}}

\author{\IEEEauthorblockN{Yuguang Yue, Chris Green, Samuel Hunt, Irakli Salia, Wenzhe Shi, Jonathan J Hunt}
\textit{Player2}\\
jj@player2.game}


\maketitle

\begin{abstract}
We introduce Pixels2Play-0.1 (\modelname), a foundation model that learns to play a wide range of 3D video games with recognizable human-like behavior. Motivated by emerging consumer and developer use cases—AI teammates, controllable NPCs, personalized live-streamers, assistive testers—we argue that an agent must rely on the same pixel stream available to players and generalize to new titles with minimal game-specific engineering. \modelname ~is trained end-to-end with behavior cloning: labeled demonstrations collected from instrumented human game-play are complemented by unlabeled public videos, to which we impute actions via an inverse-dynamics model. A decoder-only transformer with auto-regressive action output handles the large action space while remaining latency-friendly on a single consumer GPU. We report qualitative results showing competent play across simple Roblox and classic MS-DOS titles, ablations on unlabeled data, and outline the scaling and evaluation steps required to reach expert-level, text-conditioned control.
\end{abstract}


\section{Introduction}

Artificial intelligence (AI) has been applied to game playing since its inception\cite{turing1953digital}.  
Human performance in video games correlates with intelligence \cite{kokkinakis2017exploring,peters2021construction}. Games provide a cheap, safe, and quantifiable environment for evaluating new approaches.

In parallel, large language models (LLMs) such as ChatGPT\,\cite{brown2020language} have ushered general-purpose AI into daily life.  
Most commercial LLM offerings are now multi-modal visual language models (VLMs), accepting images as input. However, even with latency and cost constraints removed, state-of-the-art VLMs struggle to finish the first level of the 1996 shooter Quake\,\cite{zhang2025videogamebenchvisionlanguagemodelscomplete}.  

We are working to close this gap with Pixels2Play 0.1 (\modelname), a foundation model trained end-to-end to play any 3D title from raw pixels. Like a novice human, \modelname ~is expected to perform at a non-trivial level on unseen games without per-game engineering, improving with additional exposure.  
Crucially, our goal is to make the model text-conditioned. Generating behavior in response to prompts such as ``win using only an ax'' or ``play defensively'', allowing a richer human-AI interaction. The level of understanding required for this task is significantly higher than many, particularly reinforcement learning (RL) based approaches, which aim solely to speedrun the game.

\subsection{Uses for such a model}

In designing our model, we have benefited from the recent robotics literature \cite{ye2024latent, nvidia2025gr00t, agarwal2025cosmos}, particularly in the area of learning from unlabeled video data. Unlike in robotics, however, where the goal is often to alleviate a boring or dangerous task, humans play video games by choice for recreation. This leads to an obvious question: why would we want an AI to play video games?

We have been using early versions of our model to explore a number of consumer experiences enabled by our work.

\begin{itemize}
    \item \textbf{Gaming companions.} Co-operative titles are often more enjoyable with a friend; When friends are unavailable, our AI can keep the experience social.
    \item \textbf{Adaptive NPCs.} Pixel-driven control frees designers from brittle scripts, enabling richer, emergent interactions without game-specific code.
    \item \textbf{Play-while-you-watch assistants.} Players can let the model handle repetitive sections and intervene when desired - a personalized live streamer on demand.
    \item \textbf{Automated QA.} Text prompts let testers focus the agent on edge cases (e.g.\ stress testing collision, exploring hidden areas), accelerating bug discovery.
\end{itemize}

\section{Related work}

The literature on AI in game play is large; here we focus on recent work on video games.

One focus has been the use of reinforcement learning (RL) to train models to play games. This requires instrumenting the game to extract a reward, such as the score or win/loss of the game. For that reason, this approach is typically limited to playing a single game. Most of the results are based on model-free RL; \cite{hafner2025mastering} is a notable exception.  Many approaches also substitute engineered state representations for raw pixels, adding additional game-specific engineering.  With abundant computation, RL can achieve superhuman play,\cite{mnih2013playing,vinyals2019grandmaster,berner2019dota}, although the resulting policies often differ markedly from the human style.

Behavior cloning (BC) reframes control as supervised learning.  \cite{pearce2022counter, kanervisto2025world} applied behavior cloning in a single game, while \cite{raad2024scaling} trained a single model in multiple games. \cite{tuyls2023scaling, pearce2024scaling} have investigated the scaling laws of behavior cloning. \cite{farhang2024humanlike} used offline RL to generate human-like behavior, but in an engineered state space rather than from pixels.

\cite{baker2022video} trained a behavior cloning model to play Minecraft. Most of the training data was ``unlabeled'' from online video sources. In order to use these data, this work first trained an inverse dynamics model (IDM) from the labeled data and then used this to impute labels on the unlabeled data. Similar ideas have recently become popular in robotics, where learning from unlabeled videos is now common.  Most robotics papers generate \emph{latent} actions: an IDM infers latent controls that are trained to encode information, via a forward dynamics (world) model, useful for predicting the next frame.  Policies are first trained on those latent actions and later mapped to real motor commands \cite{ye2024latent,bjorck2025gr00t,liang2025clam}.

\section{Methods}

\subsection{Policy model}

Here we detail our architecture and training procedure for a general game-playing policy \modelname.

\begin{figure}[htbp]
    \centering
    \includegraphics[
        width=\linewidth]{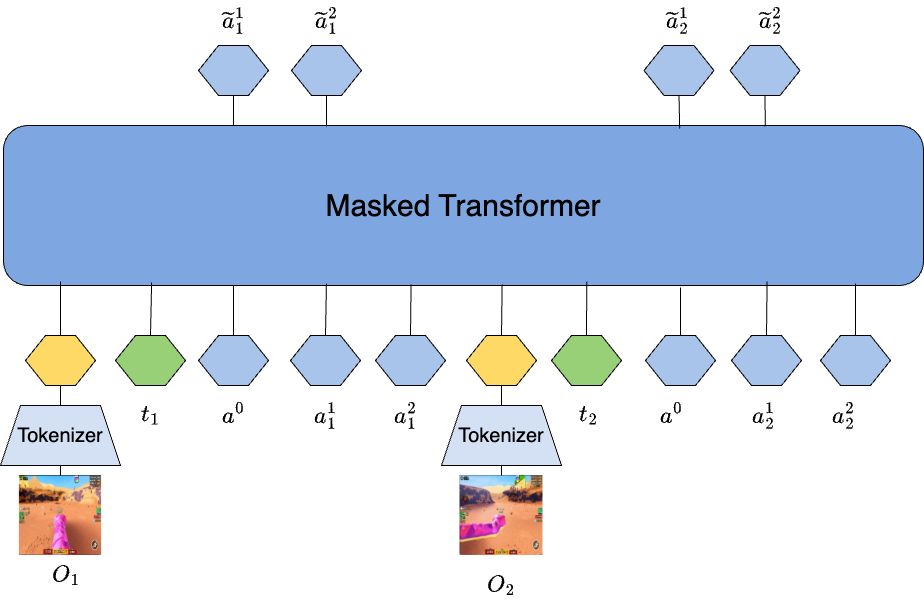}
    \caption{
    Architecture of \modelname.
Each video frame $o_{i}$ is tokenised and fed to a decoder-only transformer, followed by $k$ learnable “thinking’’ tokens $t_{i}$ that grant the model extra computation time.
The network then generates the sub-actions $a_{i}^{j}$ auto-regressively; the special token $a^{0}$ is a learnable embedding that marks the start of the action sequence for step $i$.
}
    \label{fig:arch}
\end{figure}

As in recent work, the core network is a decoder-only transformer\,\cite{vaswani2017attention} (Fig.\ \ref{fig:arch} shows the architecture).
Each video frame is first tokenized (see the tokenizer details below) and then fed to the policy transformer; we append a small number of ``thinking’’ tokens that allow extra computation before an action is emitted.  
Both training and inference run at 20\,Hz.

Actions are generated auto-regressively.  Previous single game studies collapsed the entire control vector into one categorical prediction\cite{baker2022video}.  
Our many-game setting must accommodate the full keyboard–mouse space, including up to four simultaneous key presses.  
An autoregressive factorization avoids combinatorial explosion of the action space and supports any distribution over the action space. Continuous inputs (e.g. mouse motion) are discretized, and all heads are trained with cross-entropy loss.  
The trade-off is higher inference cost, because the network is rolled forward once per sub-action; we mitigate this with standard key-value-caching and the model runs in real-time on a single RTX 5090 GPU.

\begin{figure}
    \centering
    \includegraphics[width=0.6\linewidth]{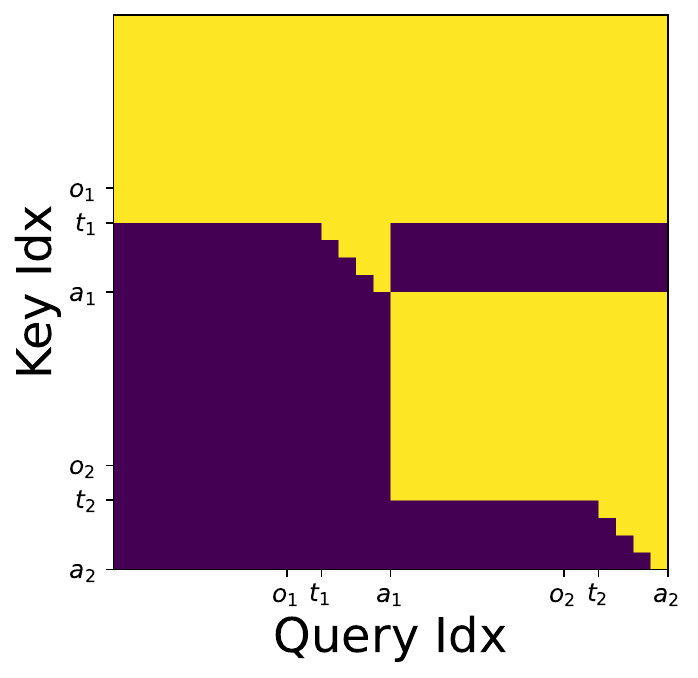}
    \caption{
    Attention mask used in our transformer policy (yellow denotes $1$ and blue $0$).
    Tick marks show the boundaries of successive inputs.  
The mask can be read by looking on the x-axis for the query and translating up to see the parts of the key masked out, e.g.\ tokens in $o_1$ can attend to any other tokens in $o_1$ and $t_1$. The ``staircases`` are the autoregressive mask while sampling the actions. Finally, you can observe in step 2 that the prior actions in step 1 are masked out.
}
    \label{fig:mask}
\end{figure}

A conventional causal mask blocks attention from future tokens. That is overly restrictive here because all image tokens arrive together in a single pass.  
Instead, we apply the causal mask only to the auto-regressive action segment, as illustrated in Fig.~\ref{fig:mask} (all well as preventing attending to future frames).

Behavior-cloning agents often suffer causal confusion \cite{de2019causal}: e.g.\ keys are often held for multiple frames, and the network may learn to copy the previous action rather than attend to pixels.  
The offline metrics looked good, but the online gameplay was poor until we masked past actions.

Masking prior actions sacrifices optimality in certain edge cases.  
For instance, in the game Need for Speed, the player may shift into gear at any point during the pre-race countdown; without seeing earlier actions, the model cannot learn the gradually increasing probability of shifting into gear.  
Empirically, the policy remains adequate. We expect to enable the action history once larger models and data sets reduce the overfitting risk.


\subsection{Inverse Dynamic Model (IDM)}

Unlabeled gameplay videos greatly outnumber curated demonstrations, so we rely on an Inverse Dynamics Model (IDM) to turn those videos into additional training data.
Two IDM approaches appear in the literature: a \emph{real-action} model that predicts explicit key/mouse actions \cite{baker2022video}, and a \emph{latent-action} model that predicts abstract action codes later mapped to real actions \cite{schmidt2023learning,ye2024latent,nvidia2025gr00t}.
For its simplicity and ability to scale, this work adopts the real-action variant; a direct comparison with latent-action IDMs is left for future study.

Formally, the IDM is a classifier over the action at time $t$ given the surrounding image sequence:
\[
\hat{a}_{t}\;\sim\;p_{\text{IDM}}\bigl(a_{t}\mid o_{1},o_{2},\dots,o_{t},\dots,o_{T}\bigr).
\]
Our IDM architecture first encodes the frame stack with a 3D convolutional block \cite{baker2022video}.  
The resulting embeddings feed a decoder-only transformer \emph{without} a causal mask, allowing the network to freely attend to past and future frames when inferring the action for each timestep.

Training minimizes cross-entropy between the predicted distribution $\hat{a}_{t}$ and the ground truth action $a_{t}$ on a labeled dataset.  
After convergence, the IDM predicts actions for unlabeled clips, yielding a much larger, imputed-labeled corpus.  We then train the policy on the full mix of labeled and imputed-labeled data.


\subsubsection{Image tokenizers}

We have tested a variety of approaches to image tokenizers including using a pre-trained (but with weights unfrozen) convolutional net \cite{pearce2022counter}, linear projections of image patches \cite{dosovitskiy2020image}, and pre-training a MagVitV2 tokenizer \cite{yu2023language} on our dataset (both unlabeled and labeled data). All three deliver broadly similar policy performance.  
We have found (as in \cite{schafer2023visual}) that using public pre-trained general image tokenizers performed poorly, probably because these tokenizers are typically trained on a large amount of photos, which differ in their visual qualities from games. We also notice that when playing a game the agent must attend to small, fast-moving cues (e.g., the small ball in Blade Ball or the miniature creature in Be a Snake), whereas standard image-recognition models often only capture broader, scene-level context.

\subsubsection{Data collection}\label{data-collection}

For unlabeled video data, we use public data sources and commercially available VLMs to curate the dataset. We implemented a two-step filtering process. First, an initial filter is applied based on metadata such as the video's title, description, topic, and thumbnail image (when available). This step involves querying a commercial VLM to assess the relevance of the videos to our specified query. Second, the full video content is processed by the VLM to segment and remove non-gameplay scenes, which will not be useful for our model training(e.g.\ introductions and some visual effects).

For labeled game playing, we use a combination of paid annotators who are requested to play specific games and are exploring the capture of gameplay (with consent) of our product users. 

We initially found a significant distributional shift between model training data and inference, which we traced to two factors: 1. During inference, no video compression takes place, but training data (for practical reasons) must be compressed. 2. The image resizing function differed between training (Python) and inference (Rust). We alleviated the distributional shift by using data augmentation to improve robustness, randomizing compression quality during video compression, and using the same resizing function between training and inference.

\subsubsection{Evaluation}

During training, we can easily observe both training and validation loss. However, offline metrics may not correspond to the online performance of the model. A significant challenge for evaluation is that we wish our model to play a large variety of games. Instrumenting even a single game for automated performance evaluation is time-consuming. For this reason, currently, we are primarily limited to a qualitative evaluation of the model's performance. This is an area that we intend to develop further.

\section{Experiments}

\subsection{General game playing}

Currently, we have focused on simple Roblox games along with some older MS-DOS titles. The Roblox platform has the advantage that, as it reduces the barriers to game design, it has a very wide variety of games. We are also learning MS-DOS games as part of a goal to use automatic evaluation in the future. In all games, we capture training data and evaluate directly on end-user computers with no instrumentation or modification to the games.

\begin{figure}
    \centering
    \includegraphics[height=0.18\linewidth]{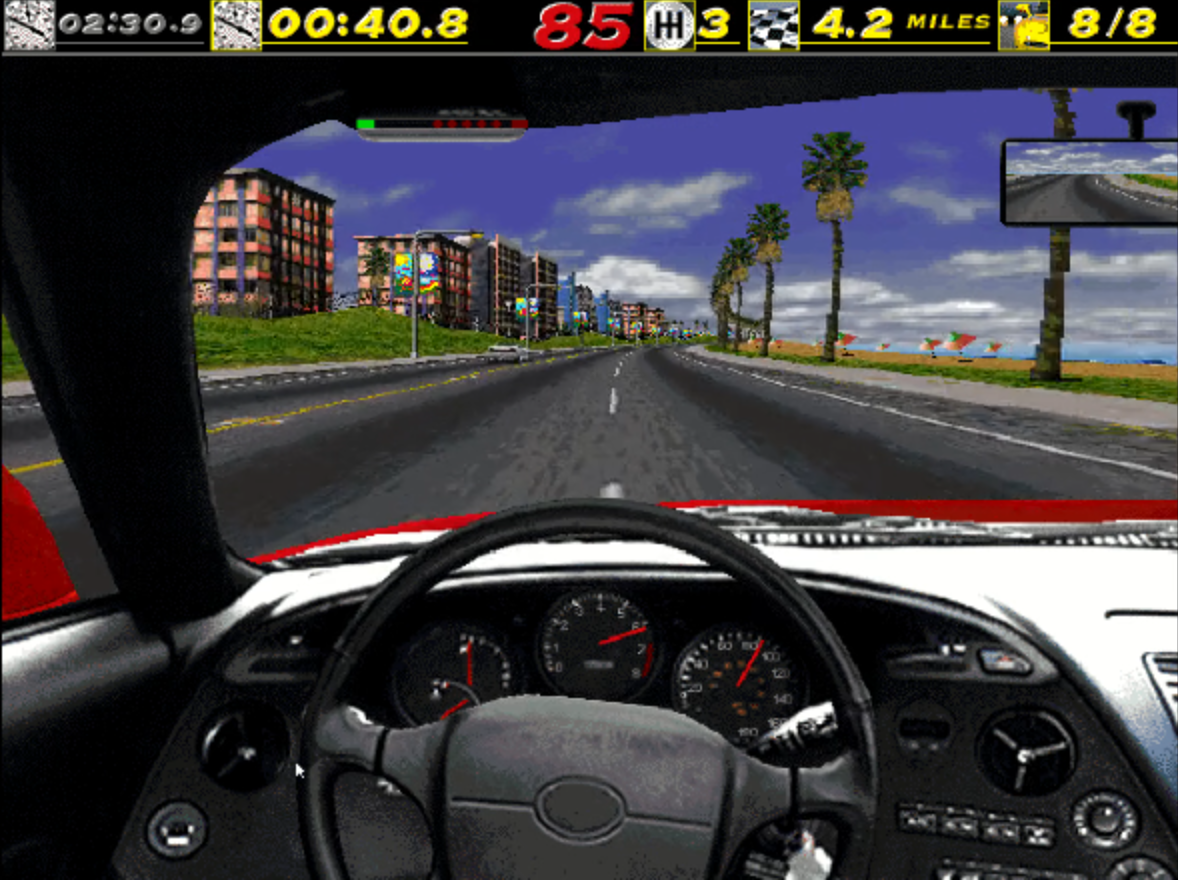}
    \includegraphics[height=0.18\linewidth]{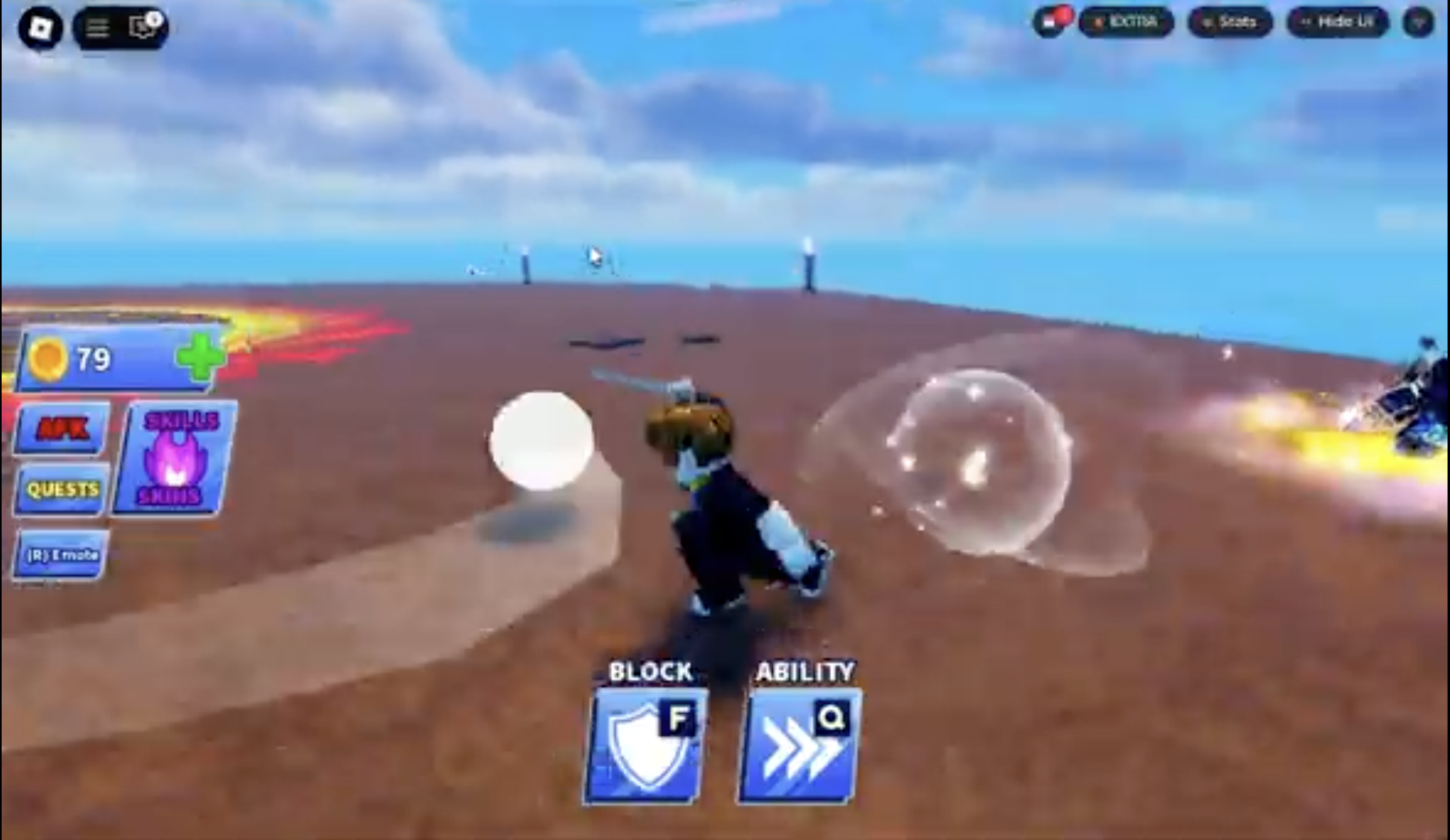}
    \includegraphics[height=0.18\linewidth]{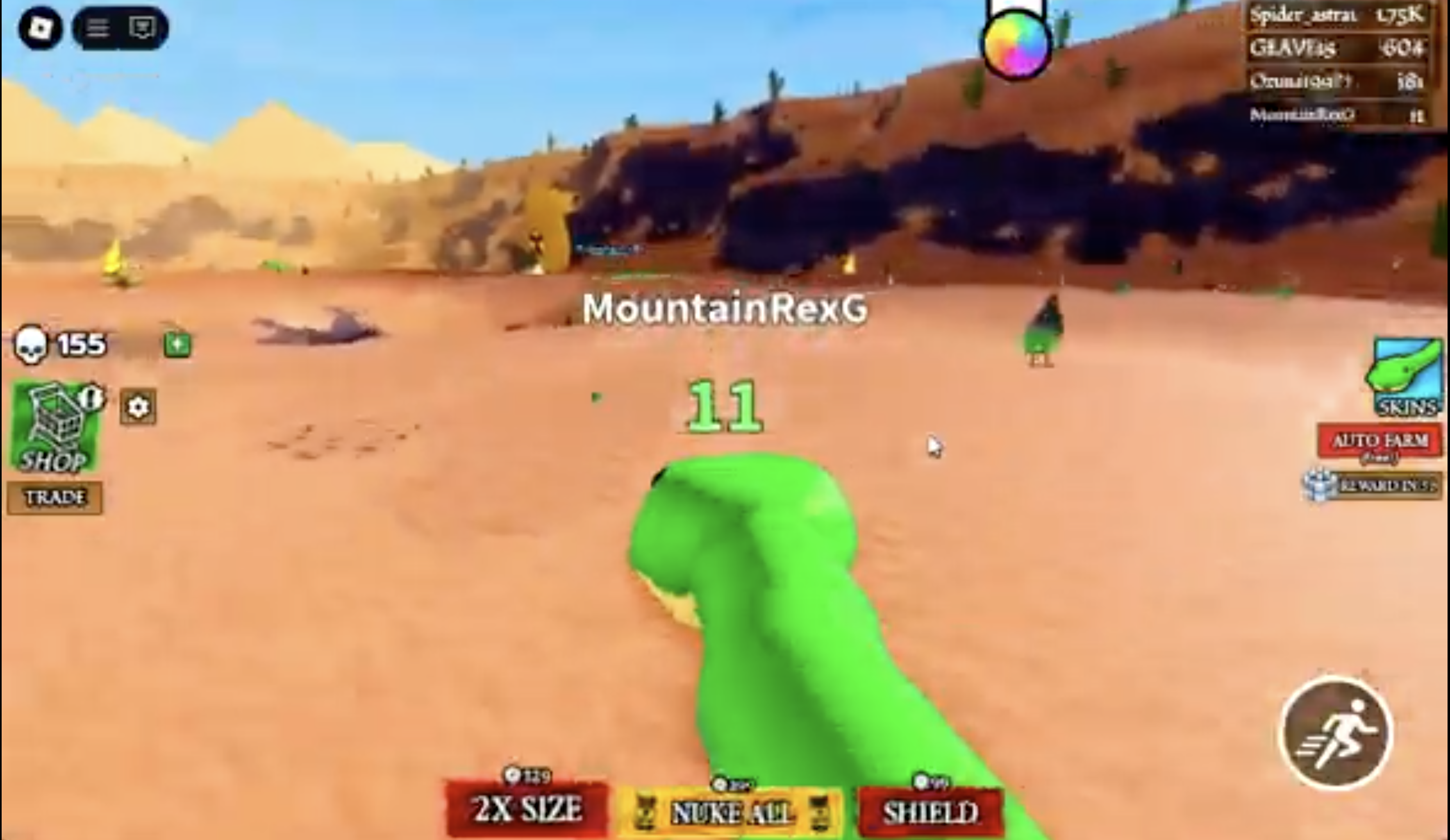}
    \caption{Examples of Roblox and MS-DOS games \modelname~is currently capable of playing. Videos of the policy in action can be viewed at the accompanying blog post \url{https://blog.player2.game/p/pixels-to-play-one-model-any-game}. We intend to make the model available for interactive demos at the conference.
    }
    \label{fig:game}
\end{figure}

As discussed above, instrumented evaluation is an area of active work. Qualitatively, we find that \modelname\ is capable of playing games currently at the level of an novice human (Fig.\ \ref{fig:game}, that is, it can play most games we trained on, but a skilled human player will outperform the model.






\subsection{Unlabeled data helps generalization} \label{unlabeled-data}
This experiment measures how additional unlabeled data affect generalization.  
The datasets follow Section~\ref{data-collection}.  
We train three variants:  
\begin{itemize}
  \item Full-Label, using 100\,\% of the training labeled data;  
  \item Limited-Label, using 10\,\% of the training labeled data;  
  \item Imputed-Label, using the same 10\,\% training labeled data plus unlabeled data being imputed by IDM.
\end{itemize}
Hyper-parameters are identical across runs, and Imputed-Label is matched to Full-Label for the total number of training frames.  
All models are validated on the same held-out labeled set.

\begin{figure}[htbp] 
    \centering 
    \centering
    \includegraphics[width=0.93\linewidth]{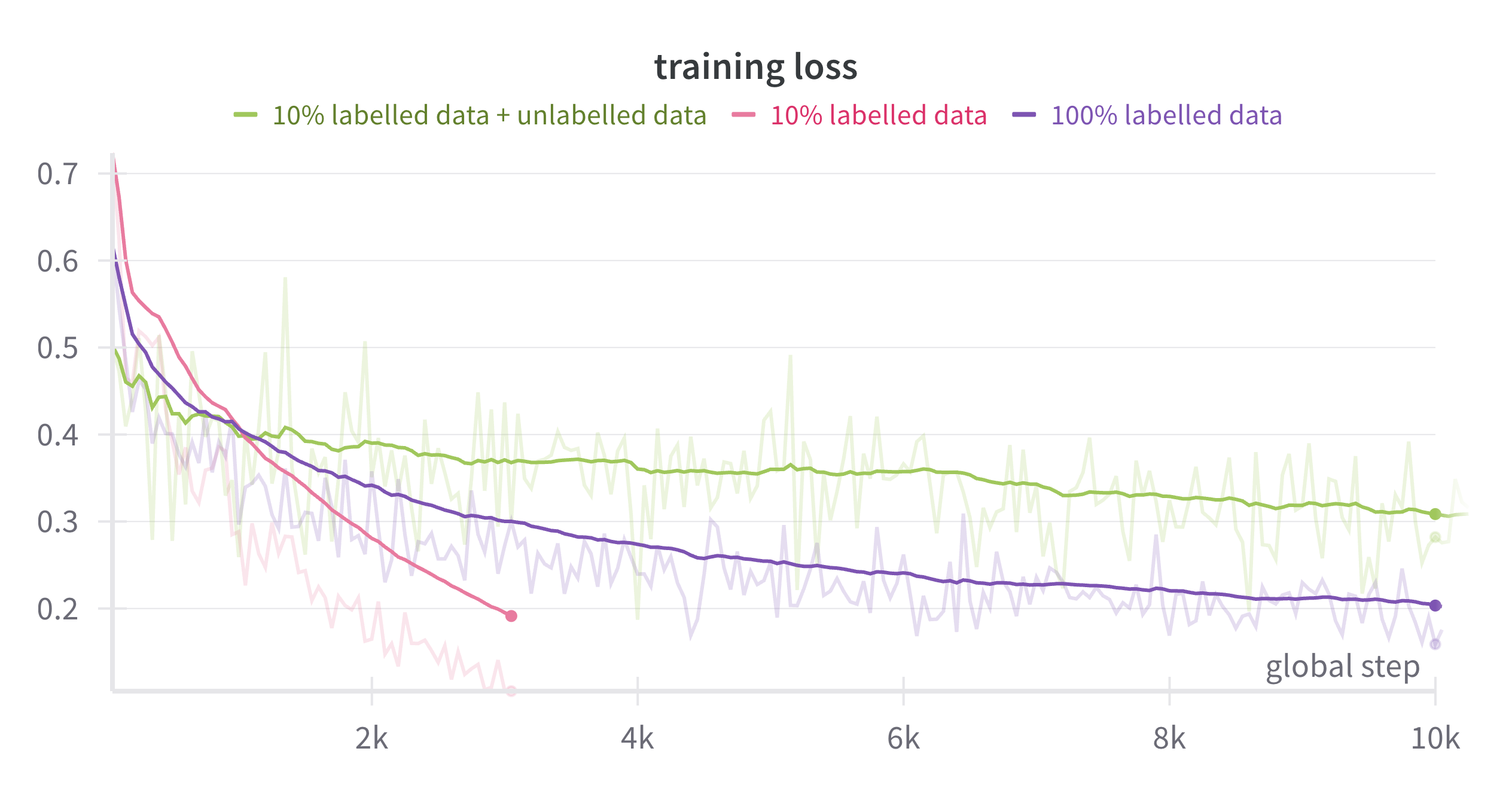} 
    \caption{Training loss curve across models with different data mixture}
    \label{training}
\end{figure}

\begin{figure}[htbp] 
    \centering 
    \centering
    \includegraphics[width=0.93\linewidth]{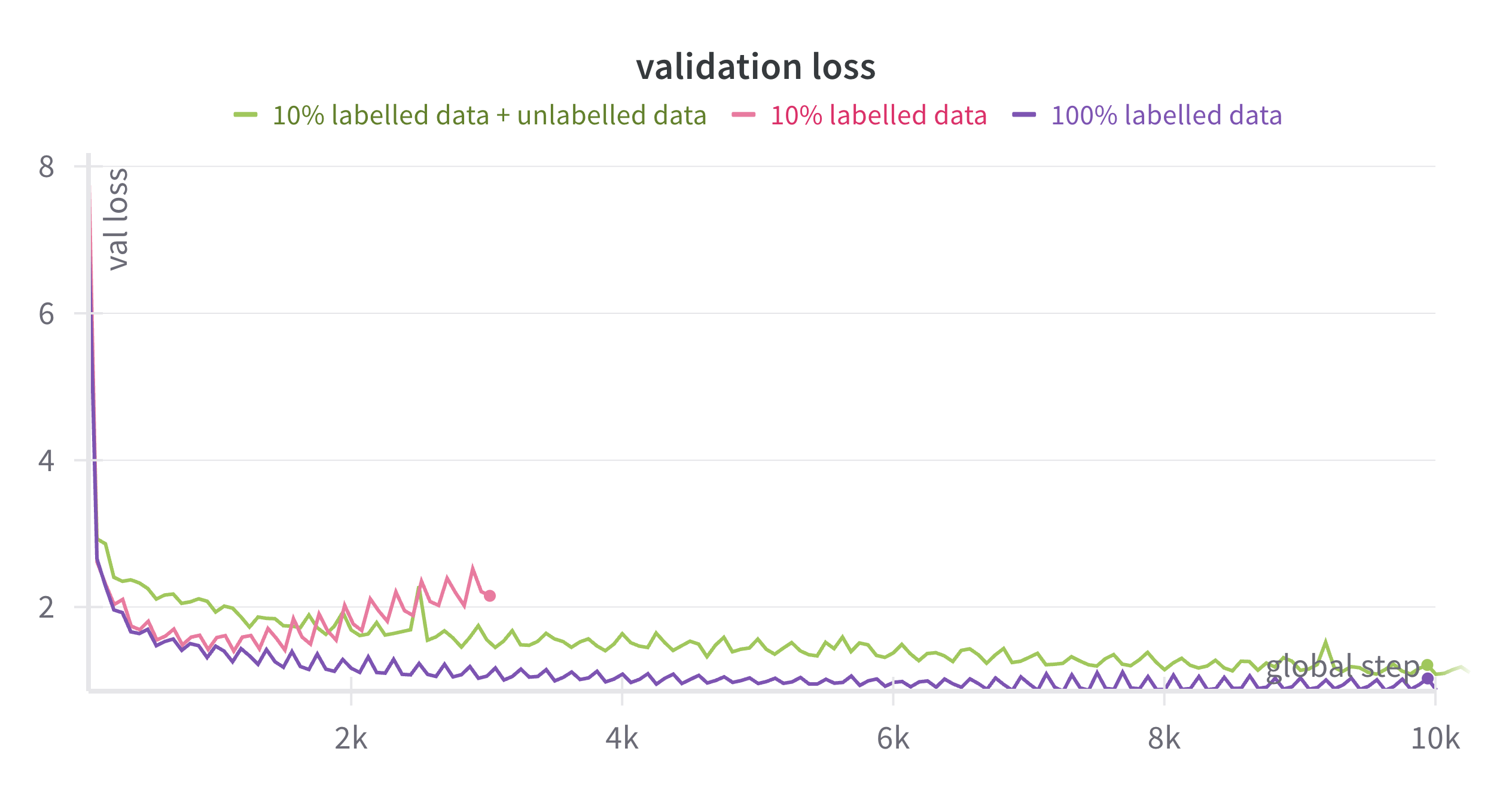} 
    \caption{Validation loss curve across models with different data mixture}
    \label{val}
\end{figure}

Figures~\ref{training} and~\ref{val} show the learning curves. Note that we stopped the run of Limited-Label training once we clearly observed overfitting to reduce the cost of resources. 
Limited-Label overfits rapidly after around $30$ epochs through the limited dataset, while Full-Label and Imputed-Label continue to improve throughout training.  
The best validation perplexities are 1.40 (Limited-Label) and 1.08 (Augmented-Label), a reduction 22\% attributable to the imputed-label data.  
A residual gap between full-label and enhanced-label highlights the quality difference and remaining domain shift between true labels and imputed labels.

\section{Discussion}

This paper reports on initial progress toward a foundation model that produces human-like behavior directly from pixels in 3D video games.  We have also discussed early user-facing prototypes—AI companions, smarter NPCs, and play-assist tools—that both validate the approach and create new streams of real gameplay data for future training.

Currently, \modelname~handles a range of relatively simple 3D titles.  The ongoing work focuses on two main fronts.  First, we continue to iterate on architecture and scaling, enlarging both the labeled and unlabeled corpora and increasing model capacity.  Second, we are extending the temporal window so that the agent can reason over much longer histories, a prerequisite for competent play in more complex games.

\bibliographystyle{IEEEtran} 
\bibliography{refs}

\end{document}